\documentclass{article}

\usepackage{PRIMEarxiv}

\usepackage[utf8]{inputenc} 
\usepackage[T1]{fontenc}    
\usepackage{hyperref}       
\usepackage{url}            
\usepackage{booktabs}       
\usepackage{amsfonts}       
\usepackage{nicefrac}       
\usepackage{microtype}      
\usepackage{lipsum}
\usepackage{fancyhdr}       
\usepackage{graphicx}       
\graphicspath{{media/}}     

\usepackage{graphicx}
\usepackage[most]{tcolorbox}
\usepackage{lipsum}
\usepackage{xcolor}
\usepackage{bm}

\usepackage{hyperref}
\usepackage{url}
\usepackage{booktabs}
\usepackage{subcaption}
\usepackage{verbatim}
\usepackage{multirow}
\usepackage{makecell}
\usepackage{siunitx}
\usepackage{arydshln} 
\usepackage[ruled,vlined]{algorithm2e}
\usepackage{amsmath}
\usepackage{cases}
\usepackage{amsmath}
\usepackage{threeparttable}
\usepackage{float}
\usepackage{natbib}
\usepackage{hyperref}

\SetKwComment{Comment}{// }{}
\DeclareMathOperator*{\argmin}{argmin}

\title{Yuan3.0 Ultra: A Trillion-Parameter Enterprise-Oriented MoE LLM

}

\author{YuanLab.ai \\ research@yuanlab.ai}

\begin{document}
\maketitle

\begin{abstract}
We introduce Yuan3.0 Ultra, an open-source Mixture-of-Experts (MoE) large language model featuring 68.8B activated parameters and 1010B total parameters, specially designed to enhance performance on enterprise scenarios tasks while maintaining competitive capabilities on general purpose tasks. We propose Layer-Adaptive Expert Pruning (LAEP) algorithm designed for the pre-training stage of MoE LLMs. In contrast to previous expert pruning approaches that operate primarily in the post-training phase, the proposed algorithm enhances training efficiency by selectively pruning underutilized experts and reorganizing experts across computing devices according to token distribution statistics. Comprehensive experiments demonstrate that LAEP effectively reduces model size and substantially improves pre-training efficiency. When pre-training Yuan3.0 Ultra from scratch original with 1515B parameters, this algorithm delivers a 49\% boost in pre-training efficiency and a 33.3\% reduction in total parameters, while preserving the model’s outstanding multi-domain performance. On enterprise scenario benchmarks including Docmatix, ChatRAG, SummEval and MMTab, Yuan3.0 Ultra achieves leading accuracy. 
The model and codes are publicly available at \href{https://github.com/Yuan-lab-LLM/Yuan3.0-Ultra}{https://github.com/Yuan-lab-LLM/Yuan3.0-Ultra}.
\begin{figure*}[h]
\centering
  \includegraphics[width=\linewidth]{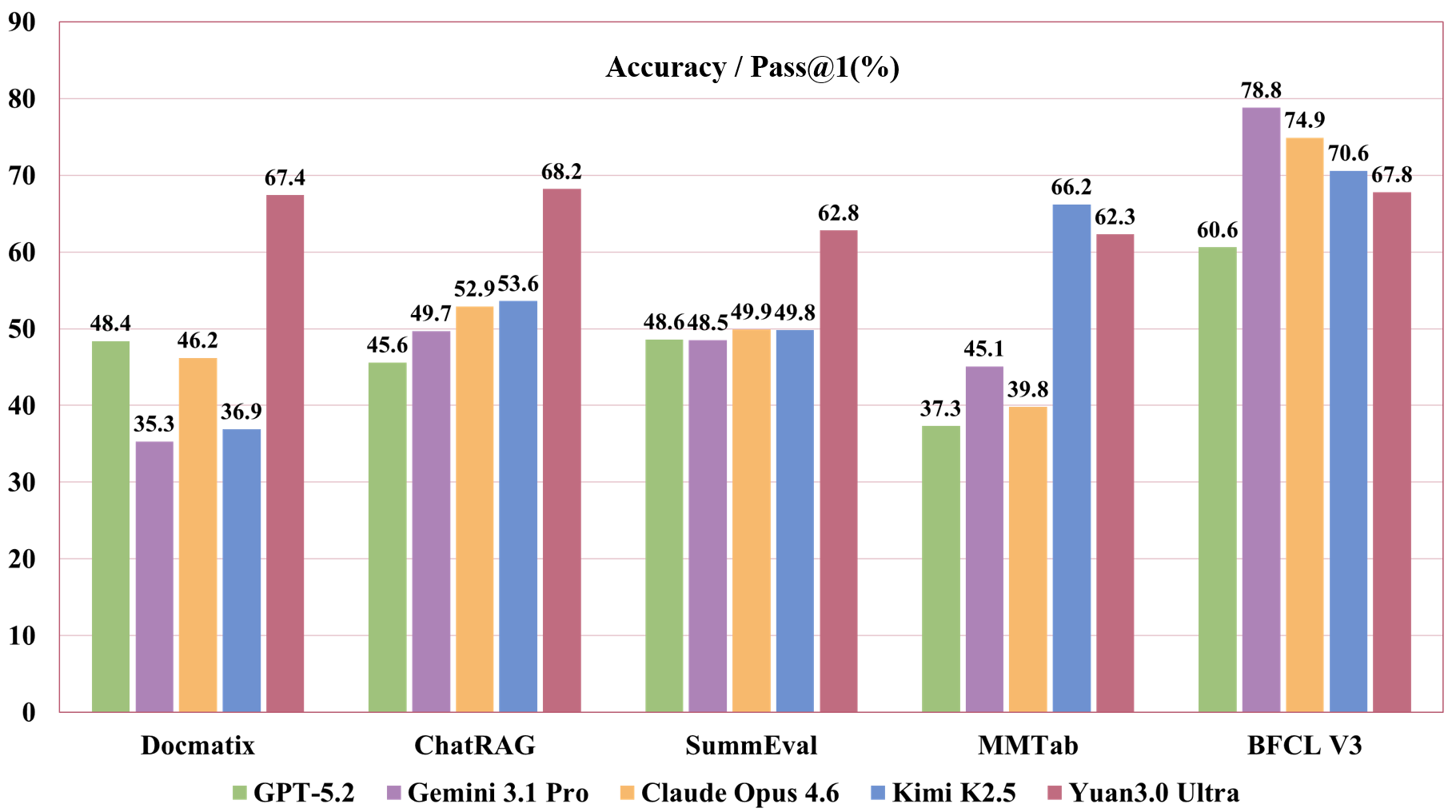}
  \caption{Performance Comparison of Yuan3.0 Ultra across enterprise scenarios benchmarks.}
  \label{fig:abstract}
\end{figure*}

\end{abstract}


\section{Introduction}

Recent advances in Mixture-of-Experts (MoE) large language models, exemplified by models like  Mixtral \citep{jiang2024mixtral}, Kimi-K2.5 \citep{kimiteam2026kimik25visualagentic}, DeepSeek-V3 \citep{liu2024deepseek}, and Qwen3.5 \citep{qwen35blog}, have achieved important breakthroughs in model scale and accuracy. 
The MoE architecture enables a significant increase in model capacity without linearly inflating computation costs (FLOPs) by activating only a small subset of total parameters for each token, though the static parameters used to construct MoE models still take up considerable memory. 
During the training process, the MoE model often experiences expert load imbalance with certain experts frequently activated while some experts are rarely engaged, which can cause some experts to struggle to learn useful representations impacting the model's overall performance and also lead to inefficient use of  computational resources.

Expert pruning is a technique that structured pruning the model by identifying and removing experts that have minimal impact on performance during training \citep{he2023structured}, thereby reducing model complexity and memory usage. Currently, research on expert pruning primarily focuses on the post-training phase \citep{liu2024efficient,lu2024not,xie2024moe}. Critically, to the best of our knowledge, no prior work has successfully applied expert pruning during the pre-training stage of MoE LLMs.

In this study, we conduct a comprehensive study on expert pruning in pre-training of large-scale MoE models. Our main contributions are summarized as follows:

\begin{itemize}
    \item Analyze the evolution of expert load throughout pre-training and identify two distinct phases. The first is an initial transition phase, spanning the beginning of training to the first several hundred iterations, during which expert loads exhibit substantial volatility inherited from random initialization, with the number of tokens routed to the same expert varying by orders of magnitude. The second is a subsequent stable phase, characterized by converged expert load and relatively minor fluctuations in per-expert token assignments.
    
    \item Propose Layer-Adaptive Expert Pruning (LAEP), a novel algorithm designed for the pre-training of MoE LLMs. Once expert loads enter the stable phase of pre-training, LAEP selectively prunes underutilized experts and rearranges the remaining experts to alleviate load imbalance across computing devices. Extensive experiments demonstrate that LAEP significantly reduces the total number of model's parameters while improving model accuracy and pre-training efficiency.
    
    \item Apply LAEP to pre-train a 1515B-parameter sparse MoE model, achieving a 33.3\% reduction in model parameters and a 49\% improvement in training efficiency compared with the base model. Moreover, the resulting Yuan3.0 Ultra pre-trained base model attains performance comparable to state-of-the-art systems on major benchmarks across diverse domains.

    \item We refine the Reflection Inhibition Reward Mechanism (RIRM), proposed in Yuan3.0 Flash \citep{wu2026yuan3}, and integrate it into a fast-thinking reinforcement learning paradigm for post-training. This enhanced RIRM yields a 16.33\% gain in training accuracy and reduces output token length by 14.38\%. As a result, Yuan3.0 Ultra achieves state-of-the-art accuracy on enterprise scenario benchmarks, 
    and remains highly competitive in various tasks covering a wide range of domains.
\end{itemize}

\section{Related Work}

\subsection{Expert Pruning for MoE models}
Current studies on expert pruning in MoE models predominantly focus on the post-training stage, where an already trained model is pruned for specific downstream tasks by retaining task-critical experts. Representative methods include structured or heuristic pruning strategies \citep{xia2023sheared,lu2024not}, evolutionary or clustering-based approaches to remove redundant experts \citep{liu2024efficient,guo2025cluster}, and one-shot pruning criteria based on weights or router-weighted activations with minimal or no retraining \citep{xie2024moe,lasby2025reap}. Recent work further identifies the existence of a small number of “super experts” whose preservation defines the effective limit of expert-level compression \citep{su2025unveiling}. Notably, these approaches rely on task- or data-specific pruning during fine-tuning or inference, in contrast to our method, which performs expert pruning directly during the pre-training phase.

\subsection{Expert load balance method for MoE models}
As MoE models scale, load balancing has become a critical concern. Prior work primarily relies on auxiliary load-balancing losses, notably introduced in the Switch Transformer \citep{fedus2022switch} and adopted by subsequent models such as Mixtral \citep{jiang2024mixtral}, to regulate expert utilization through gating probabilities. Extensions further address imbalance at token, expert, device, or sequence levels, as in DeepSpeed-MoE \citep{rajbhandari2022deepspeed}, ST-MoE \citep{zoph2022st}, and DeepSeek-V3 \citep{liu2024deepseek}. However, these auxiliary losses are highly sensitive to their weighting coefficients, often degrading perplexity and overall performance when overemphasized. In contrast, our approach eliminates auxiliary loss terms and instead, directly reorganizes token distributions across computing devices in a loss-free manner, to improve pre-training efficiency.

\section{Method}
\subsection{Expert Token distribution in Pre-training}
To investigate the load characteristics of experts during the pre-training phase of MoE LLMs, we first constructed a 20B-parameter model and pre-trained it from scratch. The complete model architecture and the details of the pre-training dataset are described in Appendix \ref{app: structure} and Appendix \ref{app: dataset}, respectively. 

\begin{figure}[t]
  \begin{center}
  \includegraphics[width=0.6\columnwidth]{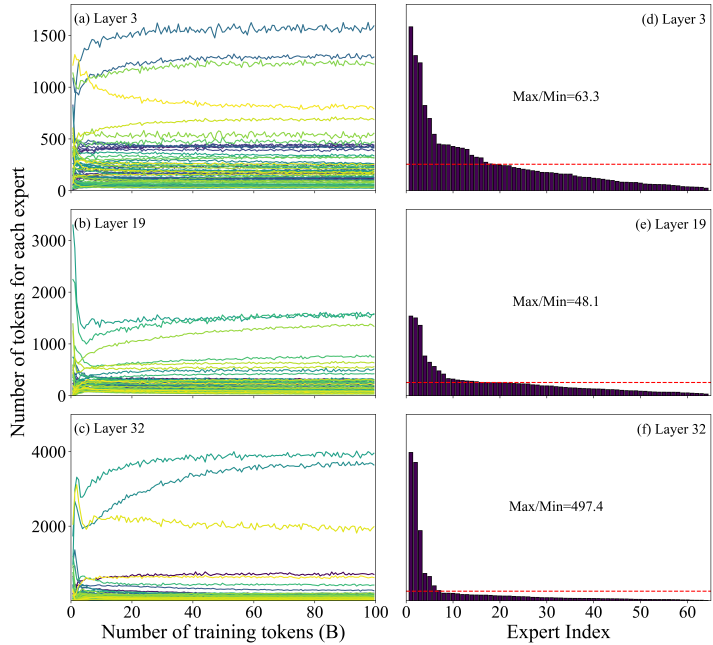}
  \end{center}
  \caption{The left column (a-c) displays the trend of token distribution among experts across different layers changing over the course of training. The right column (d-f) shows token distribution of individual experts in the stable stage of pre-training (80B tokens). The red dashed line denotes the average number of tokens per expert per layer. Max/Min represents the ratio between the token count of the most heavily utilized expert and that of the least utilized expert.   }
  \label{fig:iter}
\end{figure}

Figure \ref{fig:iter} illustrates the evolution of token distribution across experts at three representative layers of the 20B model over the course of pre-training. We observe that the expert token loads exhibits pronounced imbalance and follows a consistent temporal pattern across layers.

Overall, the evolution of expert token allocation during pre-training can be divided into two distinct phases.

\textbf{Initial transition phase}. At the beginning of pre-training, expert token load rapidly departs from the random initialization. During this phase, the distribution of tokens among experts changes drastically, and inter-expert disparities in token counts grow sharply. As training proceeds, the number of tokens assigned to each expert quickly converges toward a relatively stable regime.

\textbf{Stable phase}. 
The token distribution across experts becomes temporally stable, with per-expert token counts changing slowly over time. Despite this stability, a pronounced imbalance persists: a small subset of experts consistently processes a disproportionately large fraction of tokens, while others receive extremely few tokens and remain persistently underutilized. Without explicit load-balancing control, the disparity between the most utilized and least utilized experts can span several orders of magnitude, as shown in Figure \ref{fig:iter}. Importantly, once the stable phase is reached, the relative ordering of experts by token load becomes largely fixed. This stable ranking provides a reliable basis for identifying persistently underutilized experts.


\subsection{Expert Pruning}
The significant workload disparity among experts results in insufficient training for some experts, leading to diminished parameter efficiency and a waste of computational resources. To address this issue, we propose the LAEP method, which formulates pruning strategies adaptively at the layer level, with such strategies being determined based on the distribution of tokens across experts within each individual layer.

$E[i,j,l]$ represents whether the $j^{th} $ token in the $l^{th}$ layer is routed to the $i^{th}$ expert. 
\vspace{-6pt}
\begin{subnumcases}
    {E[i,j,l] = }
    1 & if yes\\
    0 & else
\end{subnumcases}
\vspace{-6pt}

Experts that meet the following conditions will be pruned:
\vspace{-6pt}
\begin{equation}
\label{equ:alpha}
 \sum_{j'= 1}^{S} E \left[ i,j',l \right ]\leq \frac{\alpha }{N} \sum_{i'=1}^{N}\sum_{j'=1}^{S} E\left[i,j',l\right]
\end{equation}
\vspace{-8pt}
\begin{equation}
\label{equ:beta}
  \sum_{i'= 1}^{S} E' \left[ i',l \right ]\leq \beta  \sum_{i'=1}^{N}\sum_{j'=1}^{S} E\left[i,j',l\right]
\end{equation}
\vspace{-6pt}

\begin{figure}[t]
  \begin{center}
  \includegraphics[width=1.0\linewidth]{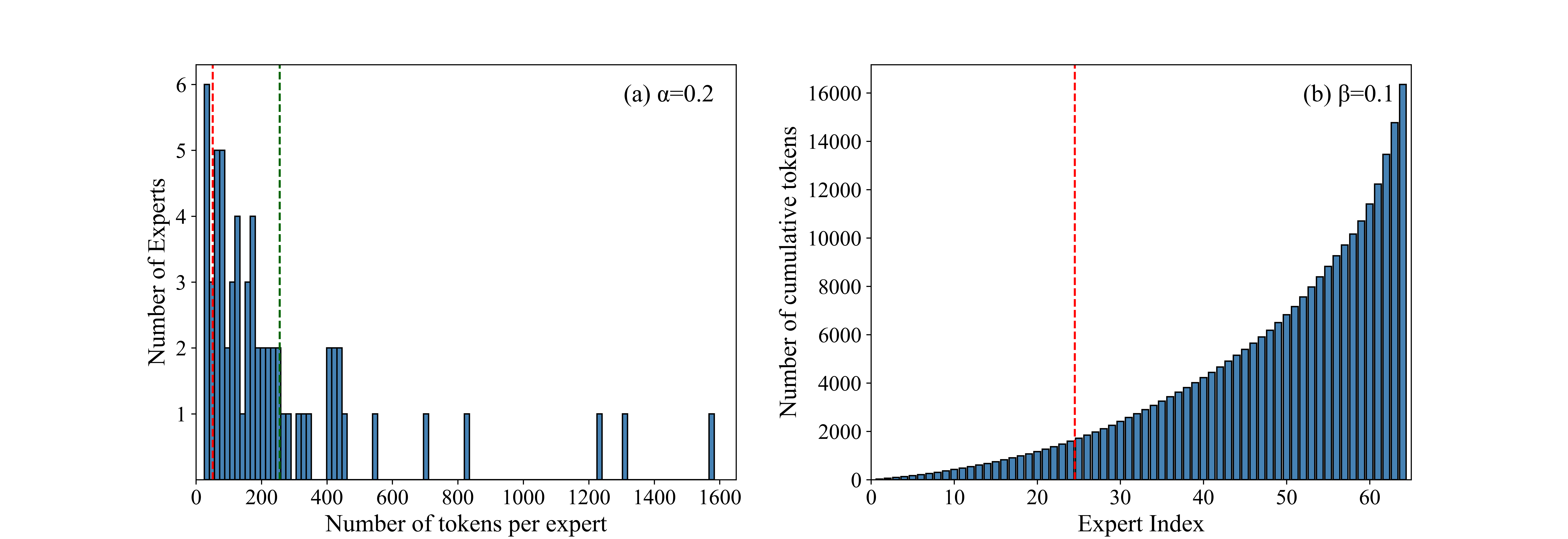}
  \end{center}
  \caption{(a) Number of experts under different token loads; (b) Accumulated tokens of experts that are sorted by token loads from smallest to largest. The area to the left of the red dashed line represents experts that need to be pruned according to hyperparameter of $\alpha$ and $\beta$, while the green dashed line represents the average token load per expert.}
  \label{fig:criteria}
\end{figure}

 $E' = \text{torch.topK}( \sum_{j'} E[i, j', l], K, \text{dim} = i, \text{largest} = \text{False} )$ is the topK experts with the smallest number of tokens among the experts in the $l^{th}$ layer. N denotes the number of experts in each layer before pruning. S is the total number of tokens per layer. $\alpha$ is the individual load constraint, and $\beta$ is the cumulative load constraint.
 Figure \ref{fig:criteria} illustrates the effects of $\alpha$ and $\beta$ on experts pruning. After sorting experts by their token loads, any expert whose cumulative token count falls below $\beta$ of the total (Equation \ref{equ:beta}) is initially flagged as candidates to be pruned, corresponding to the experts located to the left of the red dashed line in Figure \ref{fig:criteria}b. Among these flagged candidates, an expert is pruned if its individual token load is lower than $\alpha$ of the average load of all experts (Equation \ref{equ:alpha}). The detailed implementation of expert pruning algorithm is provided in Appendix \ref{app:algorithm}.

\subsection{Expert Rearrangement}
Although expert pruning effectively removes underutilized experts and mitigates extreme load disparities, significant workload imbalance across the remaining experts often persist.
This is particularly critical in large-scale pre-training, where MoE LLMs typically employ expert parallelism by distributing experts across distinct computing devices. In the framework of expert parallelism, the experts within a model are generally distributed across separate computing devices. Expert load imbalance will result in uneven computing loads across computing devices, thereby significantly diminishing the computational efficiency of the cluster.

\begin{figure*}[!htbp]
  \begin{center}
  \includegraphics[width=0.7\linewidth]{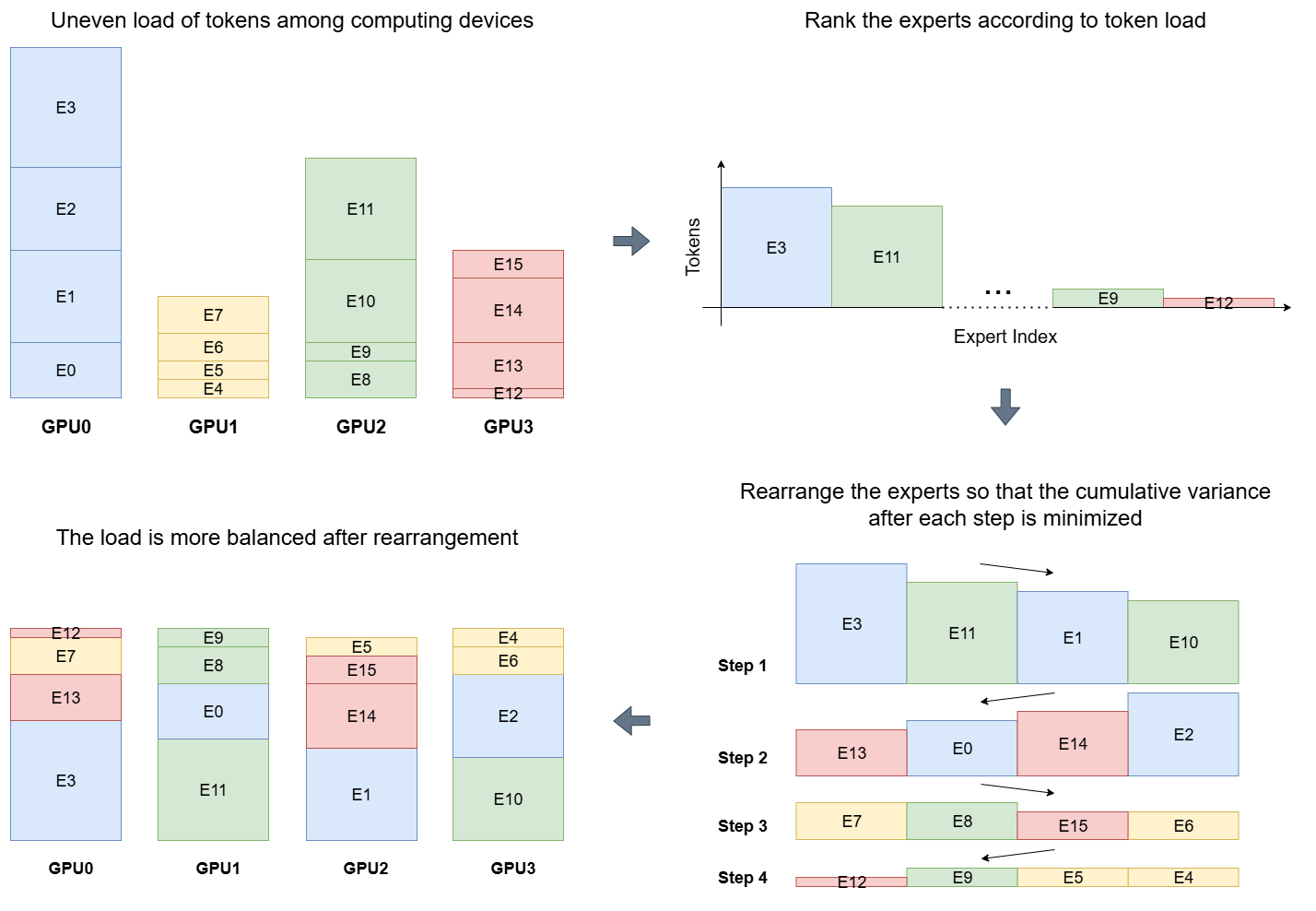}
  \end{center}
  \caption{Illustration of the Expert Rearrangement algorithm for load balancing among computing devices in MoE LLM pre-training. }
  \label{fig:balance}
\end{figure*}

To mitigate this device-level load imbalance, we propose an expert rearranging algorithm (Figure \ref{fig:balance}), which rearranges the experts across computing devices based on the expert token distribution. Starting from an imbalanced assignment of experts to computing devices, experts are first ranked according to their token loads. They are then iteratively rearranged in a greedy manner such that, at each step, the cumulative token variance across devices is minimized. By alternating the placement of high-load and low-load experts, the algorithm progressively smooths the load distribution. After rearrangement, experts are redistributed across computing devices, resulting in a significantly more balanced token load and improved utilization of computational resources. The detailed implementation of expert rearranging algorithm is provided in Algorithm \ref{alg:expert_rearranging}.

\subsection{Ablation study}

\subsubsection{Sensitivity Analysis of Pruning Hyper-parameters}
To first investigate the sensitivity of the proposed LAEP method to its pruning hyper-parameters, we analyze the effect of different pruning coefficients $\alpha$ and $\beta$ on model size and training performance.  $\alpha$ and $\beta$ exert fundamentally different influences on expert pruning. $\beta$ functions as a global pruning control parameter that emphasizing cumulative contribution of experts is low, while $\alpha$ functions as a local pruning parameter that pruning individual expert exhibiting low load compared to average load. 

Table \ref{tab:alphabeta} presents the test loss under various pruning coefficients $\alpha$ and $\beta$ on a 10B base model, the model architecture and details of the pre-training dataset are described in Appendix \ref{app: structure} and Appendix \ref{app: dataset}, respectively. When $\alpha$ is infinite, the test loss increases monotonically with $\beta$. Notably, for $\beta \leq 0.1$, the pruned model achieves a test loss of 1.658, which is lower than the base model's 1.661. This indicates that moderate pruning can effectively enhance model performance. Motivated by this observation, we fix $\beta$ at 0.1 and vary $\alpha$ to examine its impact. While the test loss exhibits an upward trend as $\alpha$ increases, it consistently remains below the baseline for $\alpha \leq 0.4$. Under this configuration, the model achieves a 24.5\% reduction in parameters compared to the base model while maintaining superior accuracy compared to base model.

\begin{table}[t] 
\caption{The impact of the hyperparameters of $\alpha$ and $\beta$ on training performance, and comparison among different auxiliary load balancing loss methods with different coefficient. }
\label{tab:alphabeta}
\begin{center}
\begin{tabular}{@{}llcc@{}}
\toprule
        &\multicolumn{1}{c}{\bf \makecell[c]{Coefficient}} & \textbf{Params (B)} &\multicolumn{1}{c}{\bf Test Loss}  \\
\midrule
Base Model &-    &9.78 & 1.661\\

\midrule

                                      &$\beta=0.05$    &8.06 &1.648\\
       Base model pruned with                &$\beta=0.1$    &6.89 &1.658\\
       varied $\beta$ and $\alpha= \infty$               &$\beta=0.2$    &5.10 &1.679\\
                       &$\beta=0.4$        &2.76 &1.725\\
\midrule
                                      &$\alpha=0.2$   &8.72 &1.643\\
    Base model pruned with                   &$\alpha=0.2$ and 0.4 hybrid   &7.89 &1.650\\
    varied $\alpha$ and $\beta=0.1$            &$\alpha=0.4$   &7.38  &1.653\\
                          &$\alpha=0.6$        &6.51  &1.661\\

\midrule
                                      &0.1  &9.78 &1.707  \\
    Base model with                   &0.01 &9.78 &1.682  \\
Deepseek-V3 Auxiliary Loss            &0.001 &9.78 &1.664   \\
                          &0.0001 &9.78 &1.656 \\
\midrule
                                      &0.1 &9.78 &1.706   \\
        Base model with                &0.01  &9.78 &1.684  \\
       Mixtral Auxiliary Loss                 &0.001 &9.78 &1.663   \\
                       &0.0001 &9.78 &1.656   \\

\bottomrule

\end{tabular}
\end{center}
\end{table}

 \begin{figure}[!htbp]
  \begin{center}
  \includegraphics[width=1.0\columnwidth]{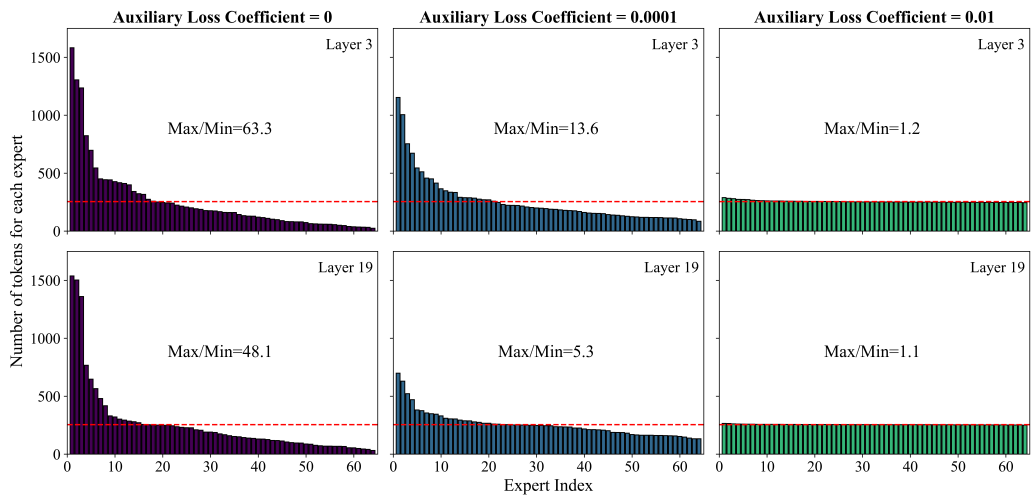}
  \end{center}
  \caption{At the stable stage of pre-training (80B tokens), the token loads of individual experts under different auxiliary loss coefficient are illustrated, with experts sorted in descending order of token count. The red dashed line denotes the average number of tokens per expert. Max/Min represents the ratio between the token count of the most heavily utilized expert and that of the least utilized expert.  }
  \label{fig:maxmin}
\end{figure}

We then examine how different auxiliary load-balancing losses influence both model accuracy and expert token distributions. Specifically, we analyze their effects on test loss (Table \ref{tab:alphabeta}) and on the distribution of tokens across experts (Figure \ref{fig:auxiliary}, Appendix \ref{app: mixtral}). We conduct experiments on the same 10B base model, comparing the baseline setting without auxiliary loss against models trained with sequence-wise auxiliary loss from DeepSeek-V3 and the auxiliary loss formulation used in Mixtral, under different weighting coefficients.

When a relatively large coefficient, such as 0.01, is applied, the auxiliary loss substantially enforces load balancing across experts as shown in Figure \ref{fig:maxmin}, but this comes at the cost of degraded model accuracy, as reflected by a higher test loss compared to the baseline. In contrast, a smaller coefficient, such as 0.0001, yields lower test loss than the baseline, while noticeable imbalance remains: in this case, the maximum token disparity within a layer can still reach approximately 13.6× (Figure \ref{fig:maxmin}).

A similar trade-off is observed for the auxiliary loss used in Mixtral. Consequently, in large-scale pre-training practice, relatively small auxiliary loss coefficients are commonly adopted (i.e., DeepSeek-V3 uses a coefficient of 0.0001 to achieve a balance between expert utilization and model accuracy). The experiment demonstrates that auxiliary loss alone cannot simultaneously achieve strong load balance and optimal performance. This limitation motivates the exploration of LAEP method, which directly leverages stable token distribution statistics to prune underutilized experts while improving model accuracy.
Compared to the auxiliary load balancing loss, LAEP not only achieves a lower test loss (1.653 vs. 1.656) but also reduces the number of model parameters by 24.5\%.

\subsubsection{Effect of pruning on different Attention Architectures}

In our base model with 10B parameters, we incorporate Localized Filtering-based Attention (LFA) to enhance local-dependency on self-attention through two successive one-dimensional convolutions \citep{wu2023yuan}. To assess whether the proposed pruning strategy generalizes to different attention variants, we remove LFA from the base model and re-evaluated the pruning behavior on self-attention. 




\begin{table}[!htbp]
\caption{Ablation study on LFA with different $\alpha$. }
\label{tab:LFA}
\begin{center}
\begin{tabular}{llcc}
\toprule
        &\multicolumn{1}{c}{\bf Pruning coefficient } &\multicolumn{1}{c}{\bf Test Loss with LFA} &\multicolumn{1}{c}{\bf Test Loss with Self-Attention}\\
\midrule
Base Model & -  & 1.661  &1.739\\
\midrule
                                      &$\alpha=0.2$ &1.643  &1.723\\
    Model pruned with                   &$\alpha=0.2$ and 0.4 hybrid &1.650 &1.729 \\
    varied $\alpha$ and $\beta=0.1$            &$\alpha=0.4$ &1.653  &1.733 \\
                          &$\alpha=0.6$      &1.661 &1.741\\

\midrule

\end{tabular}
\end{center}
\end{table}

In terms of test loss, the effect of $\alpha$ is qualitatively consistent across both settings: smaller $\alpha$ values yield lower test loss relative to the unpruned baseline, whereas increasing $\alpha$ leads to incremental increases in test loss, surpassing the unpruned level once $\alpha$ reaches 0.6 (Table \ref{tab:LFA}). However, models without LFA exhibit overall higher test loss, reflecting LFA’s contribution to improve model accuracy. These results collectively indicate that the pruning strategy remains applicable across both attention types.

\subsubsection{Impact of LAEP on Downstream Task Performance}

To validate the influence of the LAEP on the accuracy of downstream tasks, we conduct pre-training of a 20B model on a corpus of 100 billion tokens. The model’s detailed architectural specifications and the training data set are provided in Appendix \ref{app: structure} and Appendix \ref{app: dataset}.

\sisetup{
    table-number-alignment = center,
    table-format = 2.3,  
    detect-weight = true,
    detect-family = true
}

\begin{table*}[!htbp]
\caption{Comparison of LAEP and sequence-wise auxiliary loss method from DeepSeek-V3 across different tasks. }
\label{tab:perf20B}
\begin{center}
\resizebox{\textwidth}{!}{
\begin{tabular}{
    l
    S[table-format=2.3]
    S[table-format=2.3]
    S[table-format=2.3]
    S[table-format=2.3]
    S[table-format=2.3]
    S[table-format=2.3]
}
\toprule
\makecell[c]{\textbf{20B} in total \\ \textbf{0.9B} activated} &
\multicolumn{1}{c}{\bf Base Model} &
\multicolumn{1}{c}{\makecell[c]{\textbf{Base model with deepseek-v3} \\ auxiliary loss coeffcient=0.0001}} &
\multicolumn{1}{c}{ \makecell[c]{\textbf{LAEP} \\ $\alpha = 20\%$}} &
\multicolumn{1}{c}{ \makecell[c]{\textbf{LAEP} \\ $\alpha$ = 20\% and 40\% hybrid}} &
\multicolumn{1}{c}{\bf \makecell[c]{LAEP \\ $\alpha = 40\%$}} &
\multicolumn{1}{c}{\bf \makecell[c]{LAEP \\ $\alpha = 60\%$}} \\
\midrule
Total Params (B) & 19.909 & 19.909 & 19.186 & 17.776 & 16.448 & 14.624 \\
Test Loss & 1.565 & 1.563 & \textbf{1.551} & 1.556 & 1.559 & 1.567  \\
\midrule
Cmath & 17.918 & 16.872 & 20.424 & \textbf{20.778} & 19.786 & 18.556 \\
GSM8K & 10.444 & 10.083 & \textbf{10.955} & 9.742 & 9.818 & 9.685 \\
CRUXEval-O & 3.188 & 4.969 & \textbf{5.656} & 2.688 & 3.500 & 2.125 \\
TriviaQA & 7.518 & \textbf{10.204} & 8.390 & 9.354 & 8.673 & 10.091 \\
DROP & \textbf{16.568} & 15.003 & 16.384 & 16.495 & 15.026 & 15.307 \\
\addlinespace[2pt]
\cdashline{1-7}[1pt/2pt] 
\addlinespace[2pt]
\bf AVG. & 11.127 & 11.426 & \textbf{12.362} & 11.813 & 11.361 & 11.153 \\
\bottomrule
\end{tabular}
}
\end{center}
\end{table*}

Table \ref{tab:perf20B} provides a comparative analysis of the performance of the 20B models including baseline models, models integrated with sequence-wise auxiliary loss from Deepseek-V3, and the LAEP method configured 
with distinct $\alpha$ values while $\beta=0.1$. For the LAEP method, as the $\alpha$ value increases from 20\% to 60\%, the total number of parameters decrease by up to 26.5\%. In comparison to the baseline model, models pruned via the LAEP method can achieve even lower test loss than the baseline and the base model with sequence-wise auxiliary loss from Deepseek-V3.
The 20B model with $\alpha = 20\%$  produces lowest test loss and best average accuracy across five benchmarks. Notably, a more optimized pruning strategy leads to higher accuracy and 10.7\% parameters saving: when pruning the first 1/6 layers and the last 1/6 layers of the model with $\alpha = 20\%$, while setting $\alpha = 40\%$ for the middle 2/3 layers, the model shows higher accuracy with remarkably fewer parameters, compared with that of the base model and the base model with sequence-wise auxiliary loss from Deepseek-V3. No significant accuracy decline is observed across any task even when $\alpha = 60\%$.

\subsubsection{Scalability of LAEP }

We conduct scalability study of LAEP by scaling number of experts on a base model with 20B parameters.  The experimental configurations are summarized in Table \ref{tab:numexperts}. For the 128-expert setting, LAEP yields a parameter reduction of  16.3\%, and improves computational performance by 6.3\%. In the 256-expert configuration, LAEP reduces the model parameters  from 77.61B to 57.23B (a reduction of 26.3\%), accompanied by 8.3\% increase in TFLOPS. Notably, in all three cases, the LAEP delivers a better test loss compared to the base model, indicating that well-designed expert pruning not only avoids harming the accuracy, but in fact contributes to overall performance gains. Across all scales, TFLOPS improvements are substantial, which underscores LAEP’s ability to mitigate computational inefficiencies of MoE LLMs.

\begin{table}[t]
\small
\caption{Impact of LAEP on model performance by scaling number of experts.}
\label{tab:numexperts}
\begin{center}
\begin{tabular}{ccccc}
\toprule
        \multicolumn{1}{c}{\bf \makecell[c]{Num Experts}} & &\multicolumn{1}{c}{\bf Params (B)}  &\multicolumn{1}{c}{\bf TFLOPS/GPU} &\multicolumn{1}{c}{\bf Test Loss}\\
\midrule
64         &Base Model  &19.91 & 73.54  &1.565 \\
           &LAEP  &17.50 & 76.07  &1.556 \\
\midrule
128                      &Base Model &38.96 &69.82  &1.526 \\
                      &LAEP   &32.59  &74.25 &1.518  \\
\midrule
256                       &Base Model &77.61 &68.98  &1.498  \\                    
                       &LAEP  &57.23 &74.73  & 1.492  \\
                      
\midrule

\end{tabular}
\end{center}
\end{table}


\section{Pre-training of Yuan3.0 Ultra Base model with LAEP}

We applied LAEP to a 103-layer MoE model containing 1515 billion parameters.
After pruning, the number of model layers remains unaltered, whereas the total parameter count is reduced to 1010B, representing an 33.3\% reduction in parameters. This parameter optimization not only effectively alleviates the memory consumption during the pre-training phase, but also reduces the memory requirements for subsequent deployment, thereby enhancing the model's practical applicability. Details on the training data are available in Appendix \ref{app: dataset}.

\begin{table*}[t]
\footnotesize
\caption{Performance comparison of the Yuan3.0 Ultra Base model 1515B (activated 68.8B) under different methods. The tests are conducted on 824 AI chips, while the numerical precision during the training process is set to BF16. The coefficient for all auxiliary losses is set to 0.01, a value significantly higher than standard practice, to ensure near-ideal expert load balancing. Regarding the LAEP hyperparameters, $\alpha$ is set to 0.2 for the first and last one-sixth of the layers and 0.4 for the remaining layers, while $\beta$ is fixed at 0.1.}
\label{tab:perf1010B}
\begin{center}
\begin{threeparttable}
\begin{tabular}{lllll}
\toprule
\multicolumn{1}{c}{\bf Total Params}  &\multicolumn{1}{c}{\bf Expert load Balancing}  &\multicolumn{1}{c}{\bf Layers}  &\multicolumn{1}{c}{\bf Experts}  &\multicolumn{1}{c}{\bf TFLOPS}\\
\midrule
1515B  &Base model  &103 &64 &62.14\\
1515B  &Mixtral Auxiliary Load Balancing Loss  &103 &64  &80.36\\
1515B  &DeepSeek-V3 Sequence-Wise Auxiliary Loss &103  &64  &80.82 \\
\midrule
1010B    &LAEP w/o expert rearrangement &103  &48*  &82.25\\
1010B    &LAEP &103  &48*  &92.60\\
\bottomrule
\end{tabular}
    \begin{tablenotes}
            \footnotesize
            \item[*] The maximum number of experts preserved in each layer of the model
        \end{tablenotes}
        \end{threeparttable}
\end{center}
\end{table*}

As shown in Table \ref{tab:perf1010B}, the training efficiency of the model demonstrates a significant improvement after applying LAEP. Overall computational performance increases from 62.14 TFlops per GPU to 92.6 TFlops per GPU, representing a 49\% enhancement in training performance. Of these improvements, model pruning accounts for 32.4\%, while expert rearrangement contributes 15.9\%. Additionally, we compare the pre-training performance when utilizing an auxiliary loss. Notably, we employ a relatively large auxiliary loss coefficient of 0.01 to ensure near-ideal load balancing. For an analysis of the coefficient's impact on load balancing, please refer to Figure \ref{fig:maxmin}.
The incorporation of the auxiliary load-balancing loss enhances training performance by 30.1\%, which remains 12.7\% lower than the result achieved by the proposed LAEP method.

\begin{table*}[!htbp]
\caption{Comparison on tasks between the Yuan3.0 Ultra Base model and other representative dense and MoE base models.}
\label{tab:acc1010B}
\begin{center}
\begin{threeparttable}
\resizebox{\textwidth}{!}{
\begin{tabular}{llllll}
\toprule
     &\bf Benchmark  &\multicolumn{1}{c}{\bf \#Shots}  &\multicolumn{1}{c}{\bf LLaMA-3.1-405B Base}  &\multicolumn{1}{c}{\bf DeepSeek-V3-Base} &\multicolumn{1}{c}{\bf Yuan3.0 Ultra Base}\\
\midrule
          &Architecture &  &Dense &MoE &MoE\\
          &\# activated params &  &405B &37B &68.8B\\
          &\# total params  &  &405B &671B  &1010B \\
\midrule
\multirow{6}*{Language}        &Pile-test  &- &\textbf{0.54} &0.55 &0.59\\
\cdashline{2-6}[0.8pt/2pt]
\addlinespace[2pt]
          &MMLU  &1-shot &84.4 (5-shot) &\textbf{87.1} (5-shot) &78.0\\
        &ARC-Challenge &25-shot  &\textbf{95.3}  &\textbf{95.3} &94.3\\
        &NaturalQuestions  &1-shot &41.5 (5-shot) &40.0 (5-shot)  &\textbf{43.3}  \\
\midrule
\multirow{2}*{Code}  &HumanEval  &0-shot &54.9  &65.2  &\textbf{70.7} \\
      &MBPP  &3-shot &68.4 &75.4 &\textbf{75.9} \\
\midrule
\multirow{2}*{Math} &GSM8K &8-shot &83.5 &\textbf{89.3} &86.1 \\
     &MATH-500  &4-shot &49.0 &61.6 &\textbf{66.1} \\
\midrule
AVG.* &  &  &68.1 &73.4 &73.5 \\
\bottomrule
\end{tabular}
}
    \begin{tablenotes}
            \footnotesize
            \item[*] The average value after excluding Pile-test.
        \end{tablenotes}
        \end{threeparttable}

\end{center}
\end{table*}

We compare the accuracy of Yuan3.0 Ultra Base model with that of the DeepSeek-V3-Base model and the Llama-3.1-405B Base model (Table \ref{tab:acc1010B}). It can be observed that our model achieves leading accuracy on the Natural Questions, HumanEval, MBPP, and MATH-500 benchmarks. On GSM8K, the accuracy of our model is lower than that of DeepSeek-V3-Base but higher than that of Llama-3.1-405B. Additionally, on the Pile-test, MMLU, and ARC-Challenge, the accuracy of our model is lower than that of both DeepSeek-V3-Base and LLaMA-3.1-405B. Overall, the Yuan3.0 Ultra model pre-trained using the LAEP method achieves accuracy comparable to that of DeepSeek-V3-Base across various tasks, validating the effectiveness of our approach.

\section{Post Training of Yuan3.0 Ultra model}

\subsection{Training Pipeline}

Yuan3.0 Ultra employs the same multi-stage collaborative multimodal training framework as Yuan3.0 Flash \citep{wu2026yuan3}, which is elaborately designed to systematically achieve the alignment of visual and linguistic semantic spaces, the integration of cross-modal knowledge, and the progressive enhancement of complex multimodal reasoning capabilities. The overall training process of Yuan3.0 Ultra consists of four sequential stages: pretraining, multimodal pretraining, supervised fine-tuning, and large-scale reinforcement learning, with the core difference from Yuan3.0 Flash only manifested in the final reinforcement learning stage. Notably, in the final reinforcement learning phase, Yuan3.0 Ultra undergoes fast-thinking reinforcement learning training.


\subsection{Reinforcement Learning Methods}




\begin{figure*}[!htbp]
  \begin{center}
  \includegraphics[width=0.65\linewidth]{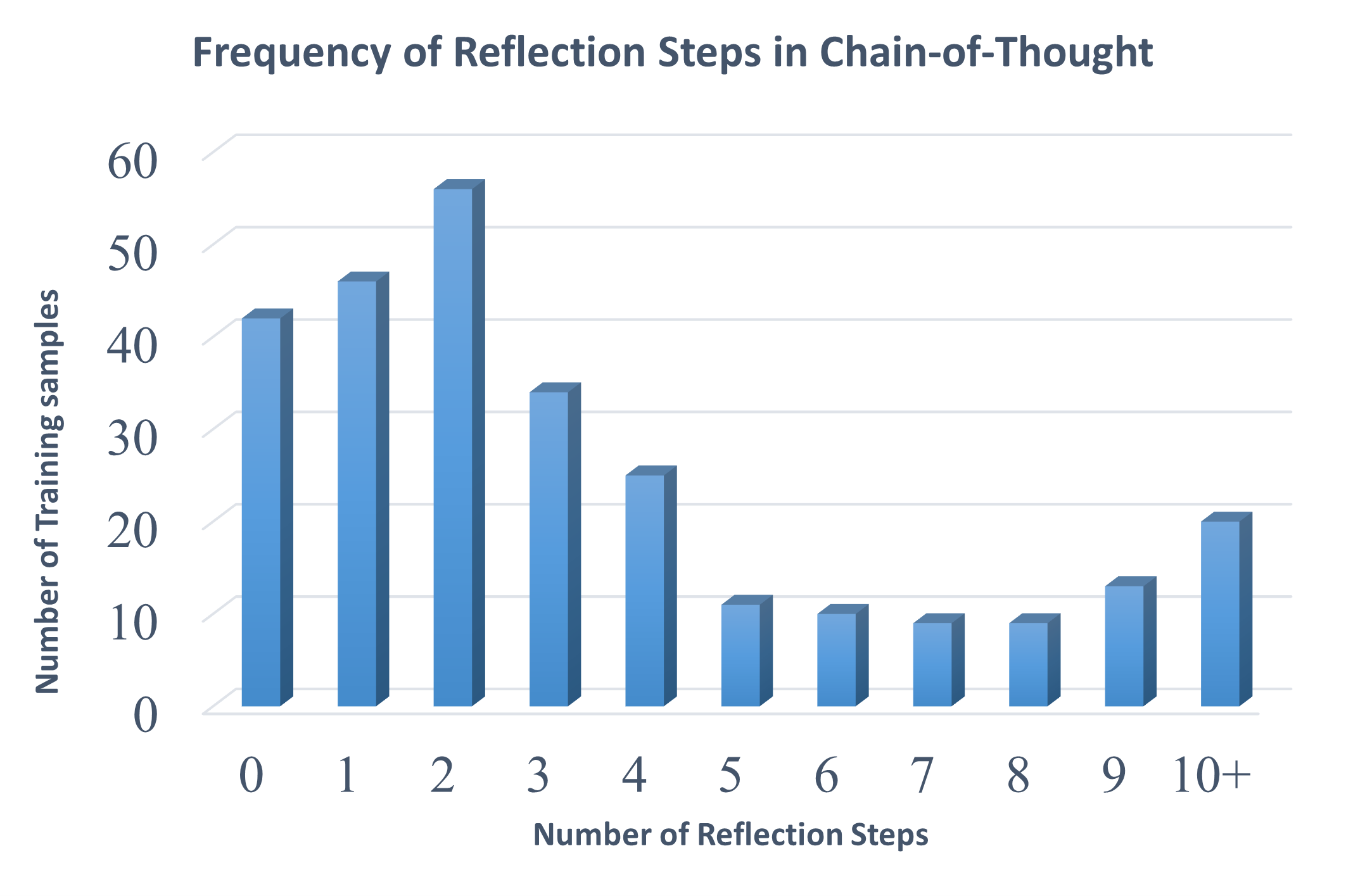}
  \end{center}
  \caption{ Frequency distribution of reflection steps in correct samples of scientific tasks during fast-thinking reinforcement learning training for Yuan3.0 Ultra.}
  \label{fig:Freqs}
\end{figure*}

In fast-thinking reinforcement learning (RL) applied to logical reasoning tasks—such as those in mathematics and science—models often produce excessively long chains of reasoning. This leads to increased computational overhead and reduced response efficiency. To address this issue, we developed the Reflection-aware Adaptive Policy Optimization (RAPO) algorithm, which has been implemented in Yuan3.0 Flash and is also employed in Yuan3.0 Ultra. Nevertheless, even under the RAPO framework, a persistent expansion of reasoning steps and prolonged output length remains evident in such tasks during fast-thinking RL of Yuan3.0 Ultra. As illustrated in Figure \ref{fig:Freqs}, many correct samples within a single training iteration present more than three reflection steps, with a non-negligible proportion exceeding ten reflection steps. To mitigate this overthinking issue, we have refined the core Reflection Inhibition Reward Mechanism (RIRM) within RAPO. 

In the revised RIRM, the calculation of $R_{ver}$ is modified as follows:
\begin{align}
    R_{ver}(v) = 
    \begin{cases}
    \max\Big(1-\frac{v-r_{min}}{r_{max}-r_{min}},0\Big)& R_{acc}>0,\\
    \max\Big(-\frac{v-r_{min}}{r_{max}-r_{min}},-1\Big)& o.w.
    \end{cases}
\end{align}
where \( r_{min}=0 \) and \( r_{max}=3 \). The reflection reward \( R_{ver} \) is dynamically computed based on two key factors: the sample accuracy score \( R_{acc} \) and the number of reflection steps \( v \). For correct samples (\( R_{acc}>0 \)), the reward increases as reflection steps approach \( r_{min} \) (ideal for fast-thinking), while for incorrect samples, more reflection steps (overthinking) trigger more severe penalties. \( r_{max} \) serves as the maximum tolerable reflection threshold: exceeding it results in zero reward for correct samples and maximum penalty for incorrect ones. This \( r_{min}=0 \) and \( r_{max}=3 \) setting aligns with fast-thinking reasoning demands: direct responses are preferred ideally, while a moderate number of reflection steps is allowed for challenging problems to ensure reasoning accuracy.

\begin{figure}[!htbp]
  \includegraphics[width=\linewidth]{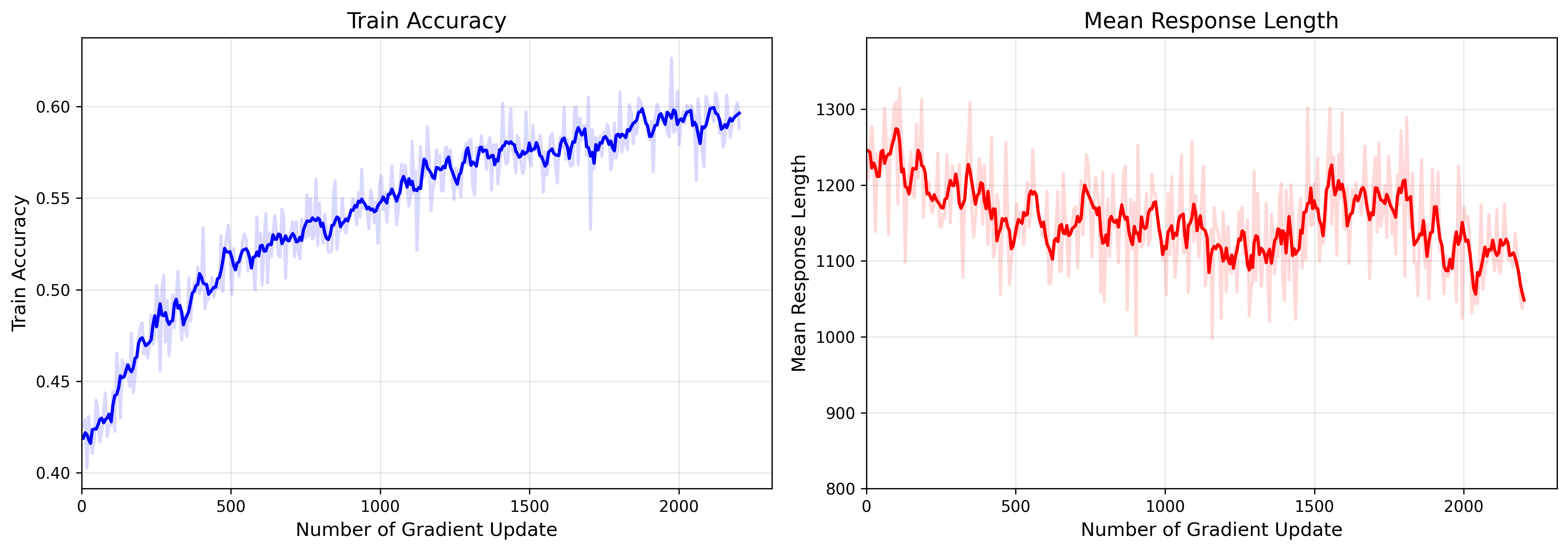}
  \caption{Training dynamics of Yuan3.0 Ultra with revised RIRM. Left: Training accuracy exhibits a steady upward trend throughout the training process. Right: Mean response length shows a gradual decreasing trend as training proceeds.}
  \label{fig:traindynamic}
\end{figure}

Figure \ref{fig:traindynamic} illustrates the training dynamics of Yuan3.0 Ultra when trained with the refined Reflection Inhibition Reward Mechanism (RIRM). The model exhibits a steady increase in reasoning accuracy alongside a concurrent, gradual decrease in its average response length throughout the training process. Specifically, this enhanced RIRM yields a 16.33\% gain in training accuracy and reduces output token length by 14.38\%. This indicates that the revised RIRM successfully promotes a consistent improvement in performance on logical and scientific reasoning tasks while effectively and strictly constraining output verbosity.

\begin{figure}[!htbp]
  \includegraphics[width=\linewidth]{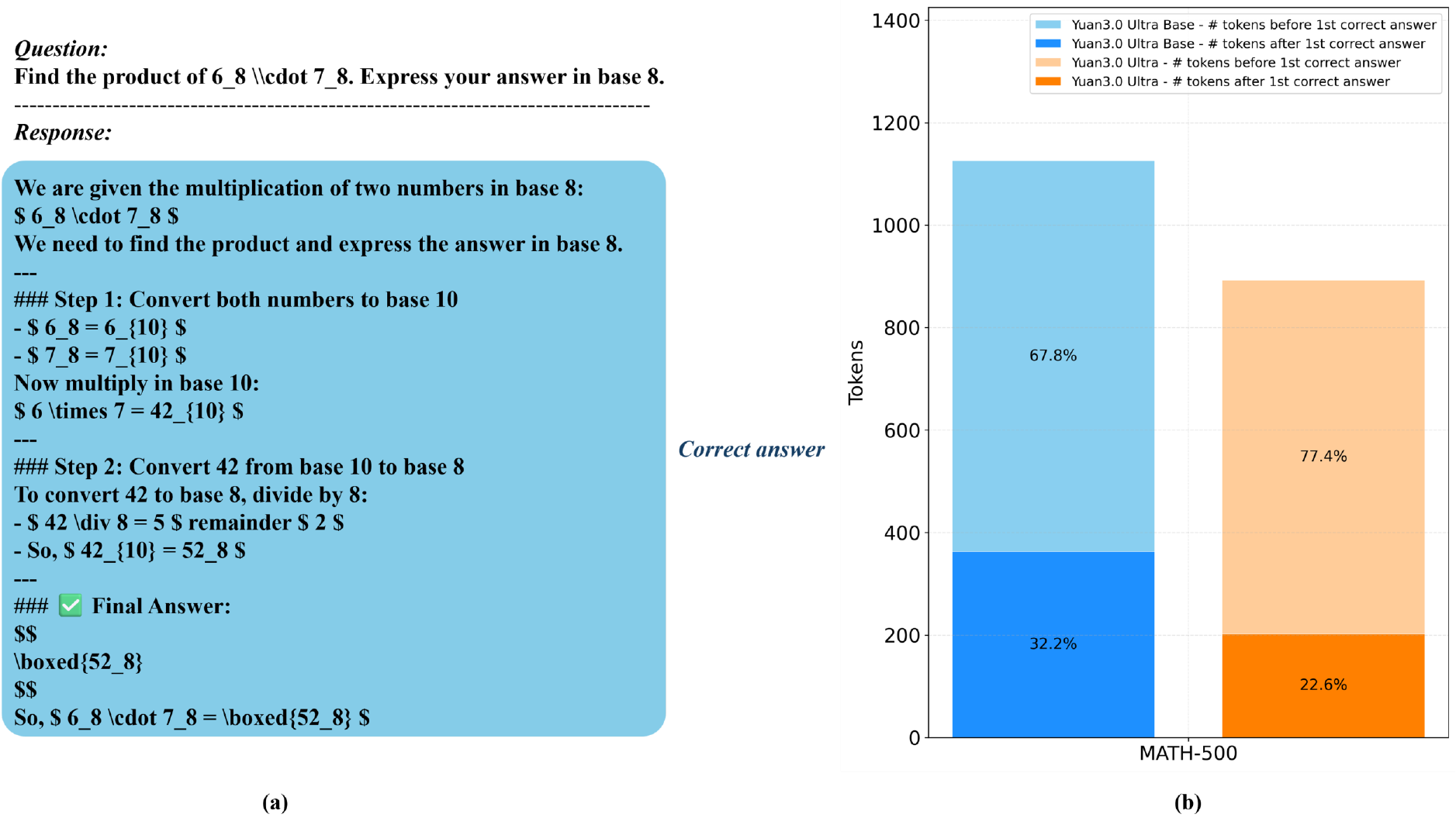}
  \caption{(a) Example of Yuan3.0 Ultra’s reasoning process after trained with revised RIRM. (b) Comparison of average token consumption of Yuan3.0 Ultra Base and Yuan3.0 Ultra on MATH-500 benchmarks. The light-colored segments indicate token consumption before the "first correct answer" appears, while the dark-colored segments correspond to token consumption during the "reflection" phase. }
  \label{fig:compare_reasoning}
\end{figure}


The integration of the revised RIRM effectively mitigates Yuan3.0 Ultra’s overthinking and optimizes its fast-thinking RL performance remarkably. As shown in Figure \ref{fig:compare_reasoning}, after training with the revised RIRM, Yuan3.0 Ultra’s reasoning process has become significantly more concise, and its token consumption throughout the reasoning process has been effectively controlled compared with Yuan3.0 Ultra Base. Quantitatively, on the MATH-500 benchmarks, Yuan3.0 Ultra achieves more efficient token usage in reasoning than Yuan3.0 Ultra Base. Notably, the token expenditure in the reflection phase, a core indicator of overthinking, has seen a substantial reduction in Yuan3.0 Ultra after the revised RIRM training, which directly validates that the revised RIRM can effectively mitigate overthinking while maintaining the model’s reasoning performance.


\subsection{Experiments}




\subsubsection{Enterprise scenario benchmarks}

We further evaluate Yuan3.0 Ultra on a suite of complex, enterprise scenario tasks, covering multimodal retrieval, text retrieval, complex table understanding, text summarization, text-to-sql, and tool invocation, which together reflect representative enterprise application requirements.

\begin{table}[!htbp]
  \centering
  \caption{Comparison among Yuan3.0 Ultra and other representative models on Docmatix}
  \begin{tabular}{lc}
    \toprule
    \textbf{Model} & \textbf{Acc(\%)} \\
    \midrule
    GPT-4o     & 56.8           \\
    o3      & 45.6            \\
   GPT-5.1     & 48.5           \\
   GPT-5.2     & 48.4          \\
   Gemini 3.1 Pro     & 35.3    \\
   Claude Opus 4.6           & 46.2     \\
   Kimi K2.5 Non-thinking & 36.9 \\
   Yuan3.0 Ultra    & \textbf{67.4}           \\
   \bottomrule
  \end{tabular}
  \label{tab:Docmatix}
\end{table}

For multimodal retrieval, we adopt the Docmatix benchmark \citep{laurenccon2024building} for multimodal RAG evaluation, which assesses a model’s ability to retrieve, associate, and accurately answer questions across multiple modalities (text, tables, and images) within multi-page, complex documents. On this benchmark, Yuan3.0 Ultra achieves the highest performance among all of the models such as Kimi K2.5, Claude Opus 4.6, Gemini 3.1 Pro and GPT-5.2, as shown is Table \ref{tab:Docmatix}. This result demonstrates strong multimodal reasoning and retrieval capabilities in document-level scenarios.

\begin{table}[!htbp]
  \centering
  \caption{Comparison among Yuan3.0 Ultra and other representative models on ChatRAG (Acc. \%)}
  \resizebox{\textwidth}{!}{
    \begin{tabular}{lccccccccccc}
        \toprule
        \textbf{Models}  &\textbf{Avg All} &\textbf{D2D}  & \textbf{QuAC}  & \textbf{QReCC}  & \textbf{CoQA} & \textbf{DoQA} & \textbf{CFQA} & \textbf{SQA}  & \textbf{TCQA}  & \textbf{HDial}  & \textbf{INSCIT}\\
        \midrule
        DeepSeek-V3    &50.5	&31.6	&28.9	&49.3	&77.0	&26.1	&83.5	&82.1	&46.7	&47.4	&32.1 \\
        GPT-4o   &50.5	&32.8	&26.6	&49.3	&76.1	&28.8	&81.9	&81.1	&49.8	&41.3	&26.7 \\                     
        o3	&44.1	&23.1	&20.8	&40.4	&69.4	&18.6	&67.8	&86.7	&45.9	&41.3	&26.7 \\          
        DeepSeek-R1	&39.4	&21.5	&22.2	&42.4	&62.5	&24.7	&81.5	&82.1	&30.7	&38.0	&28.7     \\                      
        GPT-5.1	&46.1	&28.2	&23.2	&45.4	&68.8	&20.9	&73.1	&81.3	&44.7	&45.4	&30.0     \\  
        GPT-5.2  &45.6  &30.2  &23.1  &47.0  &64.8  &25.3 &72.3  &79.1  &38.3  &45.3  &30.9      \\
        Gemini 3.1 Pro  &49.7  &33.1  &27.3  &47.0  &73.5  &34.2  &75.7  &85.5  &42.4  &48.2  &30.3  \\
        Claude Opus 4.6 &52.9  &35.3  &26.6  &49.4  &76.4  &37.3  &\textbf{86.5}  & 85.5 &50.2  &48.9  &33.2  \\
        Kimi K2.5 Non-thinking  &53.6  &34.6  &30.9  &49.9  &82.5  &35.8  &82.3  &83.6  &50.8  &51.1  &34.4  \\  
        Yuan3.0 Ultra	&\textbf{68.2}	&\textbf{55.8}	&\textbf{54.5}	&\textbf{57.3}	&\textbf{94.6}	&\textbf{63.4}	&79.8	&\textbf{91.0}	&\textbf{72.4}	&\textbf{72.9}	&\textbf{40.0}  \\             
        \bottomrule
    \end{tabular}
}
  \label{tab:chatrag}
\end{table}

For text retrieval, we evaluate on ChatRAG \citep{liu2024chatqa}, a widely recognized industrial benchmark comprising ten tasks. These tasks include long-context retrieval such as D2D, QuAC, QReCC, short-context and structured retrieval including CoQA, DoQA, CFQA, SQA, HDial, as well as Wikipedia-based retrieval tasks including TCQA, INSCIT. Across the full ChatRAG task set, Yuan3.0 Ultra attains an average accuracy of 68.2\%, surpassing Kimi K2.5, Claude Opus 4.6, Gemini Pro 3.1, GPT-5.2 and GPT-5.1, and achieves leading performance on 9 out of the 10 tasks, indicating robust retrieval effectiveness across diverse textual contexts as shown in Table \ref{tab:chatrag}.

\begin{table}[!htbp]
\centering
\setlength{\tabcolsep}{3.0pt}
\scriptsize
\caption{Comparison among Yuan3.0 Ultra and other representative models on MMTab} 
\label{tab:MMtab} 
\begin{tabular}{llcccccc}
\toprule 
 &\textbf{Tasks}  & \textbf{GPT-5.1} & \textbf{GPT-5.2} & \textbf{Gemini 3.1 Pro} & \textbf{Claude Opus 4.6} & \textbf{Kimi K2.5 Non-thinking} & \textbf{Yuan3.0 Ultra} \\
\midrule
\textbf{Avg.}   &              & 55.2  & 37.3 & 45.1 & 39.8 & \textbf{66.2} & 62.3  \\
\midrule
\multirow{5}*{\textbf{QA}}   &TABMWP (Acc.\%)      & 65.0  & 67.2 & 80.1 & 67.6 & \textbf{95.9} & 91.8  \\
                    &WTQ (Acc.\%)          & 60.8  & 69.8 & \textbf{79.6} & 76.0 & 79.3 & 77.9 \\
                    &HiTab (Acc.\%)        & \textbf{77.8}  & 15.8 & 48.3 & 44.1 & 63.9 & 67.6 \\
                    &TAT QA (Acc.\%)       & 61.4  & 28.0  & 50.5 & 44.5 & 62.4 & \textbf{74.9}  \\
                    &FeTaQA (BLEU)         & 8.7   & 6.2  & 9.6  & 12.0 & 7.4 & \textbf{39.2}  \\
\midrule
\multirow{5}*{\textbf{Fact Checking}} &TabFact (Acc.\%)      & 52.8  & 63.5 & 71.1 & 30.7 & \textbf{90.6} & 90.4  \\
                            &InfoTabs (Acc.\%)     & 64.3  & 69.3 & 74.4 & 59.6 & 81.8 & \textbf{89.7}  \\
                            &HiTab T2T (BLEU)       & 44.2  & 2.8  & 7.3  & 4.0 & 21.1 & \textbf{23.5}  \\
                            &Rotowire (BLEU)        & \textbf{17.8}  & 1.4  & 4.7 & 3.0 & 8.9 & 17.7  \\
                            &WikiBIO (BLEU)         & \textbf{12.0}  & 3.9  & 4.3 & 3.6 & 6.7 & 4.2  \\
\midrule
\multirow{2}*{\textbf{TSD}} &Row (Acc.\%)          & \textbf{96.6}  & 34.6 & 26.1 & 42.3 & 74.8 & 61.8  \\
                            &Col (Acc.\%)          & 62.1  & 82.2 & 31.7 & 68.0 & \textbf{95.2} & 81.8  \\
\midrule
\textbf{TCE} (Acc.\%)  &   & \textbf{86.4}  & 27.9 & 52.4 & 31.2 & 78.8 & 61.6  \\
\textbf{TCL} (Acc.\%)  &   & 44.7  & 17.8 & 33.0 & 43.5 & \textbf{85.2} & 69.4  \\
\textbf{MCD} (F1\%)  &    & 72.5   & 40.3 & 64.9 & 39.9 & \textbf{94.1} & 65.2  \\
\midrule
\multirow{2}*{\textbf{RCE}}  &Row (F1\%)    & 53.6  &30.2 & 51.9 & 34.6 & \textbf{83.7} & 64.0       \\
                             &Col (Acc.\%)   &57.2  & 73.5 & 76.6 & 71.9 & \textbf{94.9} & 78.0       \\
\bottomrule 
\end{tabular}
\end{table}

Multimodal table understanding is a common requirement in enterprise office scenarios. We evaluate this ability using MMTab \citep{titiya2025mmtbench}, which consists of 15 authoritative benchmarks spanning multiple task categories. Firstly, question-answering tasks include the following benchmarks. TABMWP contains multiple table-based math word problems. WTQ and HiTab are the tasks of complex querying and reasoning over Wikipedia tables. TAT-QA is a benchmark for joint reasoning over financial tables and text. FeTaQA is a benchmarks for generating free-form answer from tables. Secondly, Fact-checking task includes TabFact, InfoTabs, HiTab T2T, Rotowire, and WikiBIO. Then, TSD, TCE, TCL, MCD, and RCE focus on long-context table-centric information processing. As shown in Table \ref{tab:MMtab}, Yuan3.0 Ultra achieves a leading average accuracy of 62.3\%, exceeding Claude Opus 4.6 and Gemini 3.1 Pro, demonstrating comprehensive and well-balanced table reasoning and generation capabilities.

\begin{table*}[!htbp]
  \centering
  \caption{Comparison among Yuan3.0 Ultra and other representative models on SummEval.}
  \resizebox{1.0\textwidth}{!}{
  \begin{tabular}{l c c c c c}
    \toprule
    \textbf{Models} & \textbf{Avg. All} & \multicolumn{2}{c}{\textbf{Word Overlap}} & \textbf{Semantic} & \textbf{Factual} \\
    & \textbf{(W=100\%)} & \textbf{ROUGE-1} & \textbf{ROUGE-2} & \textbf{Similarity} & \textbf{Consistency} \\
    & & (F1\%) & (F1\%) & \textbf{BERTScore (F1\%)} & \textbf{SummaC (Acc. \%)} \\
    \midrule
    DeepSeek-V3 Non-thinking & 59.3 & 25.5 & 9.2 & 86.3 & \textbf{68.2} \\
    DeepSeek-V3.2 Non-thinking & 51.4 & 33.3 & 11.9 & 85.6 & 41.8 \\
    GPT-4o & 46.5 & 25.0 & 8.9 & 85.9 & 32.5 \\
    GPT-5.1 & 49.4 & 27.5 & 10.2 & 84.6 & 40.5 \\
    GPT-5.2 & 48.6 & 30.3 & 10.7 & 84.9 & 36.4 \\
    Gemini-3.1 Pro & 48.5 & 32.4 & 11.4 & 85.4 & 34.3 \\
    Claude Opus 4.6 & 49.9 & 33.1 & 11.0 & 85.9 & 37.8 \\
    Kimi K2.5 Non-thinking & 49.8 & 32.3 & 11.3 & 85.4 & 38.2 \\
    Yuan3.0 Ultra & \textbf{62.8} & \textbf{59.1} & \textbf{41.0} & \textbf{91.1} & 45.4 \\
    \bottomrule
  \end{tabular}
  }
  \label{tab:summeval}
\end{table*}

For text summarization, which is essential for compressing user history in agent-based applications, we evaluate on SummEval \citep{fabbri2021summeval}, covering lexical overlap, semantic similarity, and factual consistency. Yuan3.0 Ultra achieves an average score of 62.8\% as shown in Table \ref{tab:summeval}, outperforming DeepSeek-V3 (59.3\%), DeepSeek-V3.2 (51.4\%) and Kimi K2.5 (49.8\%), highlighting its strength in producing concise, semantically faithful, and factually consistent summaries.

\begin{table}[!htbp]
\centering
\caption{Performance comparison of Yuan3.0 Ultra and other representative models on Text-to-SQL benchmarks.}
\begin{tabular}{lcccc}
\toprule
\textbf{Tasks} & Qwen3.5-397B-A17B & DeepSeek-V3.2 & Kimi K2.5 & Yuan3.0 Ultra \\
\midrule
Spider 1.0        & 82.4 & 80.7 & 82.7 & \textbf{83.9} \\
BIRD              & 39.6 & 38.9 & \textbf{43.5} & 39.2 \\

\bottomrule
\end{tabular}
\label{tab:text2sql}
\end{table}

Table \ref{tab:text2sql} presents a comparative performance evaluation of the Yuan3.0 Ultra alongside other state-of-the-art models on two major Text-to-SQL benchmarks, Spider 1.0 \citep{yu2018spider} and BIRD \citep{li2023can}. The results indicate that Yuan3.0 Ultra achieves the highest execution accuracy of 83.9\% on the Spider 1.0 benchmark, slightly surpassing the performance of Kimi K2.5 and Qwen3.5-397B-A17B. On the more challenging BIRD benchmark, which emphasizes complex, real-world scenarios, 
Yuan3.0 Ultra achieves a close second with 39.2\%, marginally outperforming DeepSeek-V3.2. This comparison underscores Yuan3.0 Ultra's leading capability in structured Text-to-SQL tasks.

\begin{table*}[!htbp]
  \centering
  \caption{Comparison among Yuan3.0 Ultra and other representative models on BFCL V3.}
  \resizebox{\textwidth}{!}{
  \begin{tabular}{l c c c c c c}
    \toprule
    \textbf{Model} & \textbf{Avg. All} & \textbf{Non-Live AST} & \textbf{Live AST} & \textbf{Multi Turn} & \textbf{Relevance Detection} & \textbf{Irrelevance Detection} \\
    \midrule
    Qwen3-235B-A22B & 68.0 & 87.9 & 77.0 & 40.1 & \textbf{83.3} & 76.3 \\
    Claude-3.7-Sonnet & 58.6 & 41.3 & 78.4 & 48.4 & 72.2 & 81.4 \\
    GPT-5.2 & 60.6 & 80.9 & 76.2 & 24.6 & 72.2 & 79.7 \\
    Gemini 3.1 Pro & \textbf{78.8} & \textbf{91.5} & \textbf{84.9} & \textbf{60.3} & 61.1 & \textbf{88.2} \\
    Claude Opus 4.6 & 74.9 & 88.2 & 78.9 & 59.8 & 61.1 & 78.0 \\
    Kimi K2.5 Non-thinking & 70.6 & 86.7 & 78.6 & 48.6 & 61.1 & 77.0 \\
    Yuan3.0 Ultra & 67.8 & 81.7 & 74.5 & 45.3 & 66.7 & 86.0 \\
    \bottomrule
  \end{tabular}
  }
  \label{tab:bcfl}
\end{table*}

Finally, tool invocation ability is assessed using BFCL V3 \citep{patilberkeley}, which evaluates real-world function-calling competence across five tasks as shown in Table \ref{tab:bcfl}. Firstly,  Non-Live AST evaluates the ability of static function selection and argument extraction. Secondly, Live AST is a benchmark evaluating the ability of dynamic execution with real-time feedback. Thirdly, Multi-turn demostrates the ability of context maintenance and multi-tool coordination. Last, Relevance Detection evalutes deciding whether tool invocation is required, and Irrelevance Detection (rejecting invalid or unnecessary calls). Yuan3.0 Ultra demonstrates consistently strong performance across all categories (67.8\% on average) with no evident weaknesses, underscoring its reliability and maturity in tool-augmented reasoning and execution.

Overall, these results indicate that Yuan3.0 Ultra delivers competitive and often leading performance across a broad spectrum of enterprise-level complex tasks, with particular strengths in multimodal retrieval, table understanding, summarization quality, text-to-sql, and robust tool-calling behavior.

\subsubsection{Evaluation on Multimodal and General Reasoning Benchmarks}

\begin{table}[!htbp]
\centering
\caption{Performance comparison of Yuan3.0 Ultra and other representative models under Non-thinking mode.}
\begin{tabular}{lcccc}
\toprule
\textbf{Tasks} & Qwen3.5-397B-A17B & DeepSeek-V3.2 & Kimi K2.5 & Yuan3.0 Ultra \\
\midrule
MathVista        & 84.2 & -     & \textbf{87.4} & 76.1 \\
MATH-500         & 96.9 & \textbf{97.5} & 96.0 & 93.1 \\
HumanEval        & 97.1 & 93.1 & \textbf{97.3} & 91.4 \\
MBPP             & 91.3 & 86.8 & \textbf{94.2} & 82.0 \\
MMLU             & \textbf{91.1} & 88.6 & 90.4 & 87.8 \\
MMLU\_Pro        & \textbf{85.2} & 82.0 & 83.5 & 71.9 \\
\bottomrule
\end{tabular}
\label{tab:textnonthink}
\end{table}



We evaluate Yuan3.0 Ultra against representative large language models including Qwen3.5-397B-A17B, DeepSeek-V3.2, and Kimi K2.5 under the \emph{Non-thinking} mode. Results are summarized in Table~\ref{tab:textnonthink}.
To ensure stability on reasoning-intensive tasks, we adopt repeated sampling: 8 runs for MATH-500 \citep{lightman2023let}, HumanEval \citep{chen2021evaluating} and MBPP \citep{austin2021program}; 3 runs for MathVista \citep{lu2023mathvista}; and a single run for MMLU and MMLU-Pro \citep{wang2024mmlu}. For multi-run benchmarks, we report the average accuracy across runs. All models are evaluated under official recommended decoding settings to ensure optimal and representative performance.

Across mathematical benchmarks, Yuan3.0 Ultra achieves 
93.1\% on MATH-500, 
demonstrating solid reasoning capability. On MathVista, it obtains 76.1\%, indicating competent multimodal reasoning performance.
For code and knowledge-intensive tasks, Yuan3.0 Ultra reaches 91.4\% on HumanEval and 82.0\% on MBPP, confirming reliable code generation. 
On MMLU and MMLU-Pro, it achieves 87.8\% and 71.9\%, respectively. The results confirm competitive breadth of knowledge coverage and stable zero-shot reasoning capacity of Yuan3.0 Ultra.




\section{Conclusion}


In this paper, we introduce Yuan3.0 Ultra, an open-source Mixture-of-Experts (MoE) large language model featuring 68.8B activated parameters and 1010B total parameters, specially designed to enhance performance on enterprise scenario applications while maintaining competitive capabilities on general-purpose tasks. We propose the Layer-Adaptive Expert Pruning (LAEP) algorithm, a novel pre-training method designed for MoE LLMs that directly prunes underutilized experts to address the critical computational bottleneck of load imbalance. When pre-training the Yuan3.0 Ultra model from scratch, LAEP achieves a 33.3\% reduction in total parameters (from 1515B to 1010B) and delivers a 49\% boost in pre-training efficiency. The final Yuan3.0 Ultra model, enhanced with a refined fast-thinking reinforcement learning paradigm incorporating an improved Reflection Inhibition Reward Mechanism (RIRM), achieves state-of-the-art accuracy on enterprise scenario benchmarks, such as Docmatix, ChatRAG, and SummEval.

\section{Contribution}
Shawn Wu, Jiangang Luo, Darcy Chen, Sean Wang, Louie Li, Allen Wang, Xudong Zhao, Tong Yu, Bach Li, Joseph Shen, Gawain Ma, Jasper Jia, Marcus Mao, Claire Wang, Hunter He, Carol Wang, Zera Zhang, Jason Wang, Chonly Shen, Leo Zhang, Logan Chen, Qasim Meng, James Gong, Danied Zhao, Penn Zheng, Owen Zhu

\bibliographystyle{unsrt}  
\bibliography{references}  

@article{fedus2022switch,
  title={Switch transformers: Scaling to trillion parameter models with simple and efficient sparsity},
  author={Fedus, William and Zoph, Barret and Shazeer, Noam},
  journal={Journal of Machine Learning Research},
  volume={23},
  number={120},
  pages={1--39},
  year={2022}
}

@inproceedings{rajbhandari2022deepspeed,
  title={Deepspeed-moe: Advancing mixture-of-experts inference and training to power next-generation ai scale},
  author={Rajbhandari, Samyam and Li, Conglong and Yao, Zhewei and Zhang, Minjia and Aminabadi, Reza Yazdani and Awan, Ammar Ahmad and Rasley, Jeff and He, Yuxiong},
  booktitle={International conference on machine learning},
  pages={18332--18346},
  year={2022},
  organization={PMLR}
}

@article{zoph2022st,
  title={St-moe: Designing stable and transferable sparse expert models},
  author={Zoph, Barret and Bello, Irwan and Kumar, Sameer and Du, Nan and Huang, Yanping and Dean, Jeff and Shazeer, Noam and Fedus, William},
  journal={arXiv preprint arXiv:2202.08906},
  year={2022}
}

@article{jiang2024mixtral,
  title={Mixtral of experts},
  author={Jiang, Albert Q and Sablayrolles, Alexandre and Roux, Antoine and Mensch, Arthur and Savary, Blanche and Bamford, Chris and Chaplot, Devendra Singh and Casas, Diego de las and Hanna, Emma Bou and Bressand, Florian and others},
  journal={arXiv preprint arXiv:2401.04088},
  year={2024}
}

@article{liu2024deepseek,
  title={Deepseek-v3 technical report},
  author={Liu, Aixin and Feng, Bei and Xue, Bing and Wang, Bingxuan and Wu, Bochao and Lu, Chengda and Zhao, Chenggang and Deng, Chengqi and Zhang, Chenyu and Ruan, Chong and others},
  journal={arXiv preprint arXiv:2412.19437},
  year={2024}
}

@article{lu2024not,
  title={Not all experts are equal: Efficient expert pruning and skipping for mixture-of-experts large language models},
  author={Lu, Xudong and Liu, Qi and Xu, Yuhui and Zhou, Aojun and Huang, Siyuan and Zhang, Bo and Yan, Junchi and Li, Hongsheng},
  journal={arXiv preprint arXiv:2402.14800},
  year={2024}
}

@article{liu2024efficient,
  title={Efficient expert pruning for sparse mixture-of-experts language models: Enhancing performance and reducing inference costs},
  author={Liu, Enshu and Zhu, Junyi and Lin, Zinan and Ning, Xuefei and Blaschko, Matthew B and Yan, Shengen and Dai, Guohao and Yang, Huazhong and Wang, Yu},
  journal={arXiv preprint arXiv:2407.00945},
  year={2024}
}

@article{guo2025cluster,
  title={Cluster-Driven Expert Pruning for Mixture-of-Experts Large Language Models},
  author={Guo, Hongcheng and Yao, Juntao and Wang, Boyang and Du, Junjia and Cao, Shaosheng and Di, Donglin and Zhang, Shun and Li, Zhoujun},
  journal={arXiv preprint arXiv:2504.07807},
  year={2025}
}

@article{xie2024moe,
  title={Moe-pruner: Pruning mixture-of-experts large language model using the hints from its router},
  author={Xie, Yanyue and Zhang, Zhi and Zhou, Ding and Xie, Cong and Song, Ziang and Liu, Xin and Wang, Yanzhi and Lin, Xue and Xu, An},
  journal={arXiv preprint arXiv:2410.12013},
  year={2024}
}

@article{su2025unveiling,
  title={Unveiling Super Experts in Mixture-of-Experts Large Language Models},
  author={Su, Zunhai and Li, Qingyuan and Zhang, Hao and Qian, YuLei and Xie, Yuchen and Yuan, Kehong},
  journal={arXiv preprint arXiv:2507.23279},
  year={2025}
}

@article{he2023structured,
  title={Structured pruning for deep convolutional neural networks: A survey},
  author={He, Yang and Xiao, Lingao},
  journal={IEEE transactions on pattern analysis and machine intelligence},
  volume={46},
  number={5},
  pages={2900--2919},
  year={2023},
  publisher={IEEE}
}

@article{wu2023yuan,
  title={Yuan 2.0: A large language model with localized filtering-based attention},
  author={Wu, Shaohua and Zhao, Xudong and Wang, Shenling and Luo, Jiangang and Li, Lingjun and Chen, Xi and Zhao, Bing and Wang, Wei and Yu, Tong and Zhang, Rongguo and others},
  journal={arXiv preprint arXiv:2311.15786},
  year={2023}
}

@article{xia2023sheared,
  title={Sheared llama: Accelerating language model pre-training via structured pruning},
  author={Xia, Mengzhou and Gao, Tianyu and Zeng, Zhiyuan and Chen, Danqi},
  journal={arXiv preprint arXiv:2310.06694},
  year={2023}
}

@article{lasby2025reap,
  title={REAP the Experts: Why Pruning Prevails for One-Shot MoE compression},
  author={Lasby, Mike and Lazarevich, Ivan and Sinnadurai, Nish and Lie, Sean and Ioannou, Yani and Thangarasa, Vithursan},
  journal={arXiv preprint arXiv:2510.13999},
  year={2025}
}

@article{laurenccon2024building,
  title={Building and better understanding vision-language models: insights and future directions},
  author={Lauren{\c{c}}on, Hugo and Marafioti, Andr{\'e}s and Sanh, Victor and Tronchon, L{\'e}o},
  journal={arXiv preprint arXiv:2408.12637},
  year={2024}
}

@article{liu2024chatqa,
  title={Chatqa: Surpassing gpt-4 on conversational qa and rag},
  author={Liu, Zihan and Ping, Wei and Roy, Rajarshi and Xu, Peng and Lee, Chankyu and Shoeybi, Mohammad and Catanzaro, Bryan},
  journal={Advances in Neural Information Processing Systems},
  volume={37},
  pages={15416--15459},
  year={2024}
}

@article{titiya2025mmtbench,
  title={MMTBENCH: A Unified Benchmark for Complex Multimodal Table Reasoning},
  author={Titiya, Prasham Yatinkumar and Trivedi, Jainil and Baral, Chitta and Gupta, Vivek},
  journal={arXiv preprint arXiv:2505.21771},
  year={2025}
}

@article{fabbri2021summeval,
  title={Summeval: Re-evaluating summarization evaluation},
  author={Fabbri, Alexander R and Kry{\'s}ci{\'n}ski, Wojciech and McCann, Bryan and Xiong, Caiming and Socher, Richard and Radev, Dragomir},
  journal={Transactions of the Association for Computational Linguistics},
  volume={9},
  pages={391--409},
  year={2021},
  publisher={MIT Press One Rogers Street, Cambridge, MA 02142-1209, USA journals-info~…}
}

@inproceedings{patilberkeley,
  title={The Berkeley Function Calling Leaderboard (BFCL): From Tool Use to Agentic Evaluation of Large Language Models},
  author={Patil, Shishir G and Mao, Huanzhi and Yan, Fanjia and Ji, Charlie Cheng-Jie and Suresh, Vishnu and Stoica, Ion and Gonzalez, Joseph E},
  booktitle={Forty-second International Conference on Machine Learning},
  year={2025}
}

@article{lu2023mathvista,
  title={MathVista: evaluating mathematical reasoning in visual contexts},
  author={Lu, Pan and Ding, Xiaoyi and Wang, Lingfeng and Zhu, Zheng and He, Junxian},
  journal={arXiv preprint arXiv:2309.09979},
  year={2023}
}

@inproceedings{lightman2023let,
  title={Let's verify step by step},
  author={Lightman, Hunter and Kosaraju, Vineet and Burda, Yuri and Edwards, Harrison and Baker, Bowen and Lee, Teddy and Leike, Jan and Schulman, John and Sutskever, Ilya and Cobbe, Karl},
  booktitle={The Twelfth International Conference on Learning Representations},
  year={2023}
}

@article{chen2021evaluating,
  title={Evaluating large language models trained on code},
  author={Chen, Mark},
  journal={arXiv preprint arXiv:2107.03374},
  year={2021}
}

@article{wang2024mmlu,
  title={Mmlu-pro: A more robust and challenging multi-task language understanding benchmark},
  author={Wang, Yubo and Ma, Xueguang and Zhang, Ge and Ni, Yuansheng and Chandra, Abhranil and Guo, Shiguang and Ren, Weiming and Arulraj, Aaran and He, Xuan and Jiang, Ziyan and others},
  journal={Advances in Neural Information Processing Systems},
  volume={37},
  pages={95266--95290},
  year={2024}
}

@article{wu2026yuan3,
  title={Yuan3. 0 Flash: An Open Multimodal Large Language Model for Enterprise Applications},
  author={Wu, Shawn and Wang, Sean and Li, Louie and Chen, Darcy and Wang, Allen and Luo, Jiangang and Zhao, Xudong and Shen, Joseph and Ma, Gawain and Jia, Jasper and others},
  journal={arXiv preprint arXiv:2601.01718},
  year={2026}
}

@misc{kimiteam2026kimik25visualagentic,
      title={Kimi K2.5: Visual Agentic Intelligence}, 
      author={Kimi Team},
      year={2026},
      eprint={2602.02276},
      archivePrefix={arXiv},
      primaryClass={cs.CL},
      url={https://arxiv.org/abs/2602.02276}, 
}

@misc{qwen35blog,
    title = {Qwen3.5: Accelerating Productivity with Native Multimodal Agents},
    url = {https://qwen.ai/blog?id=qwen3.5},
    author = {Qwen Team},
    month = {February},
    year = {2026}
}

@inproceedings{yu2018spider,
  title={Spider: A large-scale human-labeled dataset for complex and cross-domain semantic parsing and text-to-sql task},
  author={Yu, Tao and Zhang, Rui and Yang, Kai and Yasunaga, Michihiro and Wang, Dongxu and Li, Zifan and Ma, James and Li, Irene and Yao, Qingning and Roman, Shanelle and others},
  booktitle={Proceedings of the 2018 conference on empirical methods in natural language processing},
  pages={3911--3921},
  year={2018}
}

@article{li2023can,
  title={Can llm already serve as a database interface? a big bench for large-scale database grounded text-to-sqls},
  author={Li, Jinyang and Hui, Binyuan and Qu, Ge and Yang, Jiaxi and Li, Binhua and Li, Bowen and Wang, Bailin and Qin, Bowen and Geng, Ruiying and Huo, Nan and others},
  journal={Advances in Neural Information Processing Systems},
  volume={36},
  pages={42330--42357},
  year={2023}
}

@article{austin2021program,
  title={Program synthesis with large language models},
  author={Austin, Jacob and Odena, Augustus and Nye, Maxwell and Bosma, Maarten and Michalewski, Henryk and Dohan, David and Jiang, Ellen and Cai, Carrie and Terry, Michael and Le, Quoc and others},
  journal={arXiv preprint arXiv:2108.07732},
  year={2021}
}

\appendix

\section{Appendix}

\subsection{Model Structure}
\label{app: structure}

\begin{table}[H]
\caption{Structural parameters of the models}
\label{tab:structpara}
\begin{center}
\begin{tabular}{lccc}
\toprule
     \bf Model Parameters   &\multicolumn{1}{c}{\bf 10B}  &\multicolumn{1}{c}{\bf \makecell[c]{20B}} &\multicolumn{1}{c}{\bf 1515B} \\
\midrule

Num of experts &64 &64  &64\\
Num of activated experts &2 &2 &2\\ 
Num of activated parameters  &0.52B  &1.18B &68.8B \\
Num of layers    &12  &48 &103\\
Hidden size  &1024 &1024  &4608\\
FFN hidden size      &4096  &2048 &16384  \\
Num of attention heads &4 &4 &36 \\
Attention hidden size &256 &256 &256 \\
Num of training tokens  &100B &100B &2200B \\

\bottomrule
\end{tabular}
\end{center}
\end{table}

\subsection{Pre-training Data}
\label{app: dataset}


Yuan3.0 Ultra adopts the identical pre-training data as Yuan3.0 Flash, which is built from a mixed corpus of pure text and image-text pairs, with core focuses on high-quality corpus curation and enhancing the model’s foundational capabilities for enterprise scenario.

\subsubsection{Textual Datasets}

The textual dataset sourced from web pages, encyclopedia entries, book excerpts, academic papers and code snippets with high linguistic rigor and structured logic. A FastText-based domain classifier is used to retain high-value enterprise-relevant domains (technology, science, opinion, knowledge) and increase industry data (finance, law, manufacturing, etc.), while a quality-scoring classifier filters out samples with a normalized quality score below 0.6 (0–1 scale). The corpus is supplemented with scientific data (practical mathematical applications, physics/engineering deduction) and enterprise-specific data (patent parsing, financial calculation). 

\subsubsection{Multimodal Datasets}

A large-scale dataset of image-text pairs is constructed to consolidate vision-language alignment and reasoning capabilities, with all sub-datasets derived from high-quality open-source corpora. It includes common scenario data, code, document, OCR, mathematical formula, charts and diagrams datasets, covering cross-domain vision-language interaction task requirements.

\subsection{Algorithm for LAEP}
\label{app:algorithm}

\begin{algorithm}[H]
\scriptsize
\caption{Expert Pruning}
\label{alg:expert_pruning}
\KwIn{\\
    \hspace*{2em}Sequence length \boldsymbol{$S$} \\
    \hspace*{2em}Number of experts \boldsymbol{$E$} \\
    \hspace*{2em}Number of experts selected per token  \boldsymbol{$topK$}\\
    \hspace*{2em}Number of training iterations in each stage  \boldsymbol{$ITER$}\\
    \hspace*{2em}Number of model layers  \boldsymbol{$L$}\\
    \hspace{2em} Pruning ratio \boldsymbol{$\alpha$}: If the token load of a particular expert is less than $\alpha$ of the average load of all experts, that expert will be pruned\\
    \hspace{2em} Pruning ratio \boldsymbol{$\beta$}: If the cumulative load of the experts that would be pruned is less than $\beta$ of the total number of tokens, all such experts will be pruned.
}

\KwOut{Pruning experts based on the index, $Exp'$}
\vspace{1em} 

\textbf{Step 1:} Recording  $ETN[ITER, L, E]$  \Comment{tokens per layer per expert per train iteration}
\vspace{1em}
\textbf{Step 2:} Getting $Exp_{dis}[ITER, L]$ \Comment{total number of pruned experts per layer}
 \hspace*{2em} and $Marker[ITER, L, E]$ \Comment{number of markers per layer per expert} 
 
\For{$iter$ in $ITER$}{
    \For{$layer$ in $L$}{
        $p \leftarrow \text{argsort}(ETN[iter, layer, :])$  \Comment{arrange experts in ascending order} 
        \For{$expert$ in $E$}{
            expert load tokens $T_{ep} \leftarrow ETN[iter, layer, p[expert]]$
            $T_{all\_ep} \leftarrow T_{all\_ep} + T_{ep}$ \Comment{add to all discarded tokens} 
            \If{$T_{all\_ep} < S * topK * \beta$ \textbf{and} $T_{ep} < S * topK * \alpha / E$}{
                $Exp_{dis}[layer] \leftarrow Exp_{dis}[layer] + 1$\;
                $Marker[layer, expert] \leftarrow Marker[layer, expert] + 1$\;
            }
        }
    }
}
\vspace{1em} 

\textbf{Step 3:} Pruning experts after token distribution of experts gets stable\;
\For{$layer$ in $L$}{
    $N_{dis\_ep} \leftarrow Exp_{dis}[layer]$ \Comment{number of pruned experts} 
    $Exp'[layer] \leftarrow \text{argsort}(Marker[layer, :], \text{order = descending})$ \Comment{index of pruned experts}
}
\vspace{1em} 

\Return{$Exp'$}\;
\end{algorithm}

\begin{algorithm}[H]
\footnotesize
\caption{Expert Rearranging}
\label{alg:expert_rearranging}
\KwIn{\\
    \hspace*{2em}Number of tokens allocated to each expert on average $D_t$ \\
    \hspace*{2em}number of groups $n_g$ \\
}
\KwOut{Reordered data $D_r$}
\vspace{1em} 

\textbf{Step 1: Initialize}\;
 $S_g \leftarrow \text{length}(D_t)\mathbin{//}n_g$  \Comment{group size}
sorted indices $p \leftarrow \text{argsort}(D_t,\ \text{order}=\text{descending})$  \Comment{sorted indices} 
$G \leftarrow \text{array}[1\ldots n_g]\ \text{of empty lists}$  \Comment{groups}
$G_{sums} \leftarrow \text{array}[1\ldots n_g]\ \text{initialized to }0$ \Comment{group sums }
$G_{indice} \leftarrow \text{array}[1\ldots n_g]\ \text{of empty lists}$  \Comment{group indices}
\vspace{1em} 

\textbf{Step 2: Allocate data to each group}\;
\For{$idx$ in $p$ }{
    $num \leftarrow D_t[idx]$  \Comment{number of tokens allocated to expert}
    \While{true}{
        $Min_g \leftarrow \argmin(G_{sums})$ \Comment{index of minimum value in group sums}
        \If{$\text{length}(G[Min_g]) < S_g$}{
            append $num$ to $G[Min_g]$\;
            append $idx$ to $G_{indice}[Min_g]$\;
            $G_{sums}[Min_g] \leftarrow G_{sums}[Min_g] + num$\;
            \textbf{break}\;
        }
        \Else{
            $G_{sums}[Min_g] \leftarrow \infty$  \Comment{Mark group as full}
        }
    }
}

\vspace{1em} 

\textbf{Step 3: Reorder data}\\
$In_{flat} \leftarrow \text{concatenate}(G_{indice}[1],\dots,G_{indice}[n_g])$ \Comment{Flattened indices}
$D_r \leftarrow [D_t[idx]\ \text{for}\ idx\ \text{in}\ In_{flat}]$ \Comment{Reordered Data}
\Return{$D_r$}\;
\end{algorithm}

\subsection{The role of auxiliary loss in expert load balancing}
\label{app: mixtral}

\begin{figure*}[!htbp]
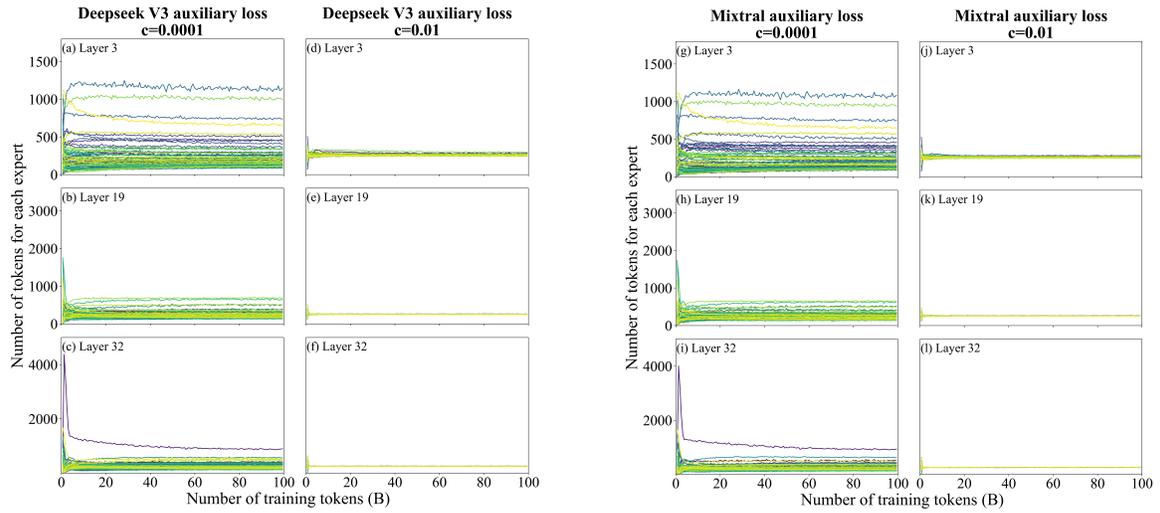


  \centering
  \begin{minipage}{0.49\linewidth}
      \centering
       \includegraphics[width=0.9\linewidth]{deepseek_app.png}
  \end{minipage}
  \begin{minipage}{0.49\linewidth}
    \centering
    \includegraphics[width=0.9\linewidth]{Mixtral_app.png}
  \end{minipage}
    \caption{The trend of token distribution among experts across different layers over the course of training. The top two columns display the results with auxiliary loss from Deepseek V3 (auxiliary loss coefficient c = 0.0001 and 0.01), while the bottom two columns show the results with Mixtral auxiliary loss (auxiliary loss coefficient c = 0.0001 and 0.01).  }
      \label{fig:auxiliary}
\end{figure*}


\end{document}